\documentclass[letterpaper, 10 pt, conference]{ieeeconf}  

\IEEEoverridecommandlockouts                              

\usepackage{amsmath}
\usepackage{bm}
\usepackage{amsfonts}
\usepackage{multirow}
\usepackage{graphicx}
\usepackage{subfig}
\usepackage{tikz}
\usepackage{cite}
\usepackage{subfig}
\usepackage{xcolor}
\usepackage{graphicx}
\usepackage{mwe}
\let\labelindent\relax 
\usepackage{enumitem}
\usepackage{adjustbox}
\usepackage[most]{tcolorbox}

\usepackage{algorithm2e}
\usepackage[font=footnotesize]{caption}
\usepackage[noend]{algpseudocode}
\usepackage{hyperref,cleveref}
\usepackage[colorinlistoftodos,prependcaption,textsize=tiny]{todonotes}
\usepackage{acronym}
\usepackage{pgfplots}
\pgfplotsset{compat=1.15}
\newlength\Fcolumnseprule
\usepackage[colorinlistoftodos]{todonotes}
\acrodef{CFAR}		[CFAR]			{Constant False-Alarm Rate}
\acrodef{FMCW}		[FMCW]			{Frequency-Modulated Continuous Wave}
\acrodef{CFEAR}	[CFEAR]	{Conservative Filtering for Efficient and Accurate Radar Odometry}

\title{\LARGE \bf
An evaluation of CFEAR Radar Odometry
}

\author{Daniel Adolfsson, Maximilian Hilger
  \thanks{
    E-mail: \texttt{dla.adolfsson@gmail.com}}
  \thanks{
    The authors are with the RNP Lab at the AASS Research Centre, \"Orebro University, Sweden}
}

\begin{document}

\maketitle
\thispagestyle{empty}
\pagestyle{empty}

\def\g5format{1}
\begin{abstract}

This article describes the method CFEAR Radar odometry, submitted to a competition at the Radar in Robotics workshop, ICRA 2024\footnote{\url{https://sites.google.com/view/radar-robotics/competition}}. CFEAR~\cite{adolfsson2021cfear} is an efficient and accurate method for spinning 2D radar odometry that generalizes well across environments. This article presents an overview of the odometry pipeline with new experiments on the public Boreas dataset. We show that a real-time capable configuration of CFEAR -- with its original parameter set in~\cite{cfear_jour} -- yields surprisingly low drift in the Boreas dataset. Additionally, we discuss an improved implementation and solving strategy that enables the most accurate configuration to run in real-time with improved robustness, reaching as low as 0.61\% translation drift at a frame rate of 68~Hz. 
A recent release of the source code is available to the community \url{https://github.com/dan11003/CFEAR_Radarodometry_code_public}, and we publish the evaluation from this article on \url{https://github.com/dan11003/cfear_2024_workshop}.
\end{abstract}

\section{Introduction}
The significance of radar in the context of autonomous systems lies in its ability to provide accurate, information-rich, and robust sensing for vehicles operating in diverse and challenging environments. This is crucial for autonomous systems as it enables safe and effective navigation, especially in scenarios where other sensors like cameras or lidar may struggle, such as in low-visibility environments. Radar does however pose specific problems due to its challenging noise characteristics. The situation has sparked research interest in how to apply radar for numerous perception tasks, including odometry estimation. and in the writing moment, motivated a competition on a public dataset.

This article describes a submission to the competition in radar odometry at the Radar in Robotics workshop at ICRA 2024. 
The competition requires methods to incrementally in real-time without utilizing future information. Competitors are ranked according to drift (i.e. translation error $[\%]$ between  100 and 800~m) in the hold-out sequences of the Boreas dataset~\cite{burnett_ijrr23}. 

\begin{figure}[htbp]
    \centering    \includegraphics[width=0.8\hsize]{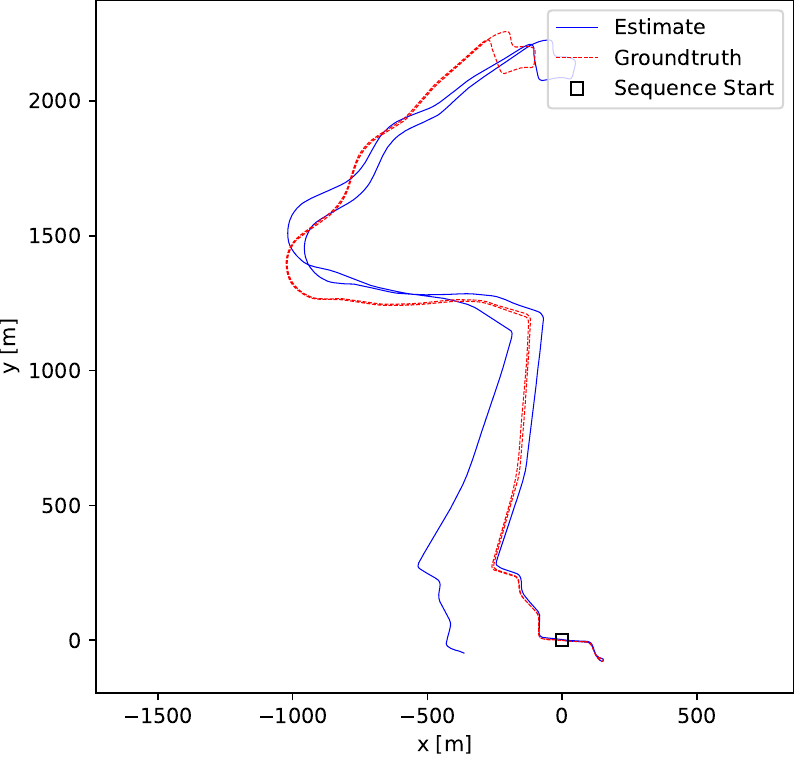}
    \caption{Estimated odometry in the snowy Boreas sequence 2021-01-26-11. Our work is demonstrated here: \url{https://youtu.be/sfX-F3xCZ_Y} 
    }
    \label{fig:trajectory}
\end{figure}

Recent work on radar odometry~\cite{lisus2024doppler,kubelka2023need} has investigated the benefit of fusing IMU or Doppler information, which can be particularly useful in geometrically degenerate cases. In contrast, the odometry pipeline presented here, namely CFEAR, operates on range-intensity measurements without additional sensors. An estimated trajectory with the method presented in this submission is presented in Fig.~\ref{fig:trajectory}.

\section{CFEAR Radarodometry}
\begin{figure}
\vspace{0.2cm}
    \centering
\includegraphics[width=0.99\hsize,trim={0cm 0cm 0cm 0cm},clip]{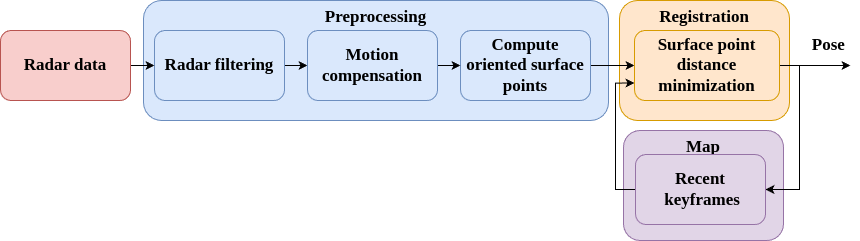}
    \caption{Overview of CFEAR Radar odometry}
    \label{fig:overview}
    \vspace{-0.5cm}
\end{figure}

The method \ac{CFEAR} is thoroughly described in~\cite{cfear_jour}, and briefly summarized here. The pipeline is designed for efficiency by extracting and matching sparse feature sets and by relying on a motion-prior to reduce the solution space. Robustness is obtained by facilitating multiple techniques for outlier rejection.
An overview is presented in Fig.~\ref{fig:overview}.

\paragraph{Feature extraction}
The input data of a $360^\circ$ sweep is given as a matrix $Z_{N_a\times N_r}$ with $N_a$ azimuths bins that each contain a total of $N_r$ range-intensity measurements.
Features are extracted using a two-step approach. First, the top k range-intensity measurements exceeding an intensity threshold $z_{min}$ are extracted per azimuth. This gives a rapid but conservative filtering of the scan. Our previous work\cite{alhashimi2021bfarbounded} gives a higher level of landmark granularity if required.

After filtering -- followed by compensating measurements for motion distortion -- features are extracted using a grid approach.
In each occupied grid cell, the intensity-weighted centroid and sample covariance is computed. This results in a sparse but descriptive representation of the scene.
\paragraph{Odometry estimation}
Motion is incrementally estimated by finding the pose $\mathbf{x}^t \in SE(2)$ that minimizes: 
\begin{equation}
\label{eq:scan_to_multikeyframes}
\begin{split}
 f_{s2mk}(\mathbf{\mathcal{M}}^{\mathcal{K}},\mathbf{\mathcal{M}}^t,\mathbf{x}^t) =\sum_{k\in\mathcal{K}}\sum_{\forall \{i,j\}\in\mathbf{\mathcal{C}}} w_{i,j}\mathcal{L}_\delta(g(m^k_j, m^t_i,\mathbf{x}^t)).
 \end{split}
\end{equation}
$\mathcal{M}^\mathcal{K}$ and $\mathcal{M}^t$  are the previously accumulated and the current set of surface points. $g$ computes the distance between neighboring surface points after applying the transformation (obtained from $\mathbf{x}^t$) to each surface point in $\mathcal{M}^t$. Correspondence sets $\mathcal{C}$ are computed between $\mathcal{M}^t$ and each keyframe $k\in\mathcal{K}$. Keyframes are created and stored in a fixed-size queue once the distance to the previous keyframe is sufficiently large.
Outliers are rejected using a robust loss function $\mathcal{L}_\delta$, i.e. Huber loss, and by weighting residuals with $w$ according to a similarity heuristic. A starting point for the optimization is provided using a constant velocity model, and the problem is solved using Levenberg–Marquardt.

A significant change to the previous implementation is that we replace the k-d tree with a hash table for rapid neighboring surface point lookup. This enables the deployment of CFEAR with significantly more keyframes in real-time.

\section{Coarse-to-fine minimization}
In rare occasions, we found that the odometry would degrade during rapid turns, resulting in inconsistencies. We hypothesize that these failures are caused by the initial guess of the optimization being outside the convergence basin. For that reason, we employ a coarse-to-fine strategy that both reduces the radius of association and increases outlier rejection in later iterations.
Specifically, a convex Huber loss is used in the first two iterations to ensure initial convergence, while a Cauchy loss is used for the remaining iterations until the termination criterion has been reached as in ~\cite{cfear_jour}.

\section{Configurations}
We use the configuration \textit{CFEAR-3} with Point-to-distribution minimization as described in~\cite{cfear_jour}, but with the filtering strategy from \textit{CFEAR-2} for higher efficiency. Apart from these rearrangements, we do not change any parameters. I.e., the parameters are manually tuned for 8 sequences in the Oxford dataset, without any tuning on the Boreas dataset. The following configurations were evaluated:
\begin{enumerate}
    \item \textit{CFEAR-3} from our previous work, 4 keyframes,
    \item \textit{CFEAR-CTF}, (1) with coarse-to-fine registration,
    \item \textit{CFEAR-CTF-S10}, (2) with keyframes increased to 10.
\end{enumerate}

\section{Evaluation}
All experiments were carried out on a Ryzen 7 7800x3d @ 5~Ghz. The pipeline runs on a single thread, with up to 8 sequence evaluations executed in parallel.
Table~\ref{tab:eval} presents the evaluation in the odometry training and test sequences of the Boreas dataset. To evaluate how the odometry generalizes across environments, we include experiments in the Oxford~\cite{RadarRobotCarDatasetICRA2020} and MulRan~\cite{gskim-2020-mulran} dataset. The same 8 respective 9 sequences as in~\cite{cfear_jour} are selected.

Surprisingly CFEAR-CTF-S10 reached as low as 0.66\% in the Boreas training set, and 0.61\% in the test set with ground truth held out. In this experiment, the coarse-to-fine strategy yields only a minor improvement (roughly 5\%) which we attribute to the otherwise rare failures with the used parameters.
Larger improvements were observed when experimenting with other parameter sets. The major improvement is attributed to additional keyframes which is feasible with the more efficient nearest neighbor search.

An estimated trajectory is shown in Fig.~\ref{fig:trajectory}, where same sequence as in ~\cite{burnett2021radar} was selected for comparison.  Interestingly, by visually inspecting the full set of estimated trajectories we observe that errors are systematic over trajectories -- i.e. the estimated end-positions are located similar to \ref{fig:trajectory}.

\begin{table}[h]

\caption{Evaluation for training and test sequences in the Boreas dataset, and 8 sequences from the Oxford training set. Drift is measured by the translation error [\%], rotation error (deg/100m). In Boreas, CFEAR-3 runs at 130.5~Hz, CFEAR-CTF at 119.8~Hz, and CFEAR-CTF-S10 at 68.0~Hz. 
}
\label{tab:eval}
\begin{tabular}{|l|l|l|l|l|l|l|}
\hline
& CFEAR-3~\cite{cfear_jour}  & CTF \textbf{(ours)} & CTF-S10 \textbf{(ours)} \\
\hline
Dataset  &  \multicolumn{3}{c|}{(transl. error [\%] / rot. error [deg/100m])}  \\
\hline
\textbf{Boreas training set}  & (0.83/0.35) & (0.79/0.35) & (0.66/0.34) \\
\hline
\textbf{Boreas test set} & - & - & (0.61/0.2)\\
\hline
\textbf{Oxford~\cite{RadarRobotCarDatasetICRA2020} seqs.~\cite{cfear_jour}} & (1.25/0.47) & (1.26/0.47) & (1.16/0.45)\\
\textbf{MulRan~\cite{gskim-2020-mulran} seqs.~\cite{cfear_jour}} &(1.29/0.39) & (1.26/0.40) & (1.18/0.38)\\
\hline
                                  
\end{tabular}
\end{table}

\section{Conclusion}
This work presented CFEAR, and discussed both robustness and efficiency improvements. An evaluation was carried out on the Boreas dataset, reaching as low as 0.61 \% drift in the test set with ground truth held out. Additional experiments in the Oxford, and MulRan dataset demonstrate a high level of generalization, with 1.16\% and 1.18\% drift respectively -- without tuning any parameters. Future work ought to investigate the cause of systematic trajectory errors.

\addtolength{\textheight}{-5cm}   
\bibliography{icra_2024_references.bib}

\end{document}